\documentclass[runningheads]{llncs}

\usepackage{eccv}

\usepackage{eccvabbrv}   
\usepackage{graphicx}
\usepackage{booktabs}
\usepackage{amsmath}
\usepackage{amssymb}
\usepackage[accsupp]{axessibility}  
\usepackage{xcolor}  

\usepackage{hyperref}
\usepackage{orcidlink}
\usepackage{rotating}

\newif\ifdraft \drafttrue

\newcommand{\ourdata}{our benthic dataset}

\begin{document}

\title{Leveraging existing sparse point annotations for benthic imagery dense segmentation}
\titlerunning{Leveraging sparse point annotations for benthic imagery dense segmentation}

\author{%
  Cesar Borja\inst{1}\orcidlink{0009-0001-2478-5573} \and
  Breck A. McCollum\inst{2}\orcidlink{0009-0003-3538-5641} \and
  Jarrett E. Byrnes\inst{3}\orcidlink{0000-0002-9791-9472} \and
  Kenneth P. Sebens\inst{4}\orcidlink{0000-0003-3445-7933} \and
  Ana C. Murillo\inst{1}\orcidlink{0000-0002-7580-9037}%
}

\authorrunning{C. Borja et al.}
\institute{%
  DIIS - I3A, Universidad de Zaragoza, Zaragoza, Spain \\
  \email{\{cborja,acm\}@unizar.es}
  \and
  Berklee College of Music, Boston, USA \\
  \email{bmccollum@berklee.edu}
  \and
  University of Massachusetts: Boston, Boston, USA \\
  \email{Jarrett.Byrnes@umb.edu}
  \and
  University of Washington, Seattle, USA \\
  \email{sebens@uw.edu}%
}

\maketitle

\begin{abstract}
The health of marine ecosystems is a critical indicator of global environmental change, yet the physical constraints of underwater observation and the intrinsic challenges of processing marine imagery severely limit the scalability of systematic monitoring. 
While recent visual foundation models such as the Segment Anything Model (SAM) series show great promise, they still struggle with the fine-grained recognition required in these complex scenarios and still require expert supervision. 
Our  work 
addresses this gap by bridging state-of-the-art foundation models with existing sparse supervision. Because historical benthic surveys are typically annotated with only a few sparse expert points per image, we utilize these legacy point-labels as visual prompts for SAM2. Our primary contribution is a novel mechanism to automatically identify which of these points are suitable, and which are actively harmful, when used for propagation. By filtering out unreliable points, we extract high-quality pseudo-ground-truth masks capable of training more accurate, fine-grained semantic segmentation models. We demonstrate the effectiveness of our approach on public benthic data and introduce a new, challenging benchmark  featuring real-world sparse expert annotations, paving the way for scalable ecological analysis.
\keywords{Automated benthic image analysis \and Fine-grained semantic segmentation \and Point-label propagation \and Pseudo-label refinement}
\end{abstract}
\section{Introduction}
\label{sec:intro}

Human-induced climate change has affected the distribution of numerous terrestrial, marine, and aquatic species around the globe~\cite{burrows2011pace,lawlor2024mechanisms}. 
During the last 60 years, approximately 84\% of all heat produced by global warming has been absorbed by the oceans~\cite{barnett2005penetration,cheng2025record}, leading to
range shifts in marine species as waters warm~\cite{sunday2015species}. 
To document and monitor these ecological range shifts, marine ecologists heavily rely on fixed-location photographic surveys of benthic environments. These visual ``snapshots'' capture crucial data on species composition over time~\cite{dunstan2007howfar}, allowing researchers to count, measure, and identify species in the laboratory rather than enduring physically demanding and logistically expensive field conditions. 

While photographic data analysis is more efficient than many methods of collecting biodiversity data in the field, the process typically followed by ecology researchers to analyze these photos is very time consuming and requires a skilled researcher with extensive expertise in species identification. 
This huge manual effort is already frequently supported by the use of software tools to assist with annotation and visualization tasks. 
Broadly used platforms like CoralNet~\cite{beijbom2012automated} have enabled a vast amount of annotations for benthic imagery databases. However, they do not contain dense, pixel-wise annotations, but rather rely on sparse point-level annotations. These  annotations typically consist of tens to a few hundred randomly or uniformly sampled pixels per image.



\begin{figure}[!tb]
    \centering
    \begin{subfigure}{0.32\linewidth}
        \includegraphics[width=\linewidth]{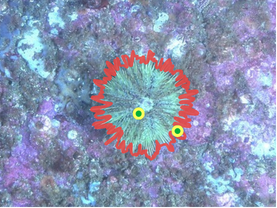}
        \caption*{\footnotesize (a) Ideal mask}
    \end{subfigure}
    \begin{subfigure}{0.32\linewidth}
        \includegraphics[width=\linewidth]{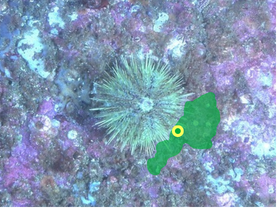}
        \caption*{\footnotesize (b) Unsuitable point-label}
    \end{subfigure}
    \begin{subfigure}{0.32\linewidth}
        \includegraphics[width=\linewidth]{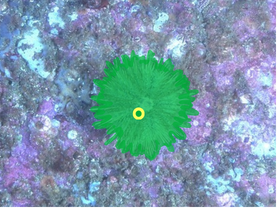}
        \caption*{\footnotesize (c) Suitable point-label}
    \end{subfigure}
    \caption{Expanding sparse expert point-labels into dense masks is a common strategy to enable automated benthic image analysis. Since sparse point-label locations were usually placed random or uniformly, they often fall on ambiguous locations. Our framework automatically identifies and filters unsuitable points to significantly improve downstream segmentation performance. a) Ideal segmentation mask for class \textit{droebachiensis} with two point-labels of the same class. b) \textit{Unsuitable} point-label whose propagation bleeds across class boundaries, introducing severe label noise. c) \textit{Suitable} point-label that propagates correctly.}
    \label{fig:intro}
\end{figure}

Recent advances in computer vision, particularly visual foundation models, offer a promising pathway to automate further benthic image analysis via semantic segmentation. 
However, even the top performing models, such as the SAM model family~\cite{kirillov2023sam, ravi2024sam2, carion2025sam}, sill struggle with fine grained and rare classes segmentation, precisely the case for most of the classes included in these datasets.  
Therefore, further supervision from experts to obtain more accurate and specific models. Since often the expert supervision is only sparse point labels, a common paradigm to train dense segmentation models on existing benthic data is \textit{point-label propagation}. In this process, sparse expert point-labels are expanded into dense pseudo-ground-truth masks, leveraging superpixels~\cite{alonso2019coralseg,raine2022point} or more recently deep features like DINOv2~\cite{raine2024human}.

However, a critical flaw in current pipelines is the implicit assumption that all points are equally suitable for propagation. Because ecological survey points are historically sampled randomly or uniformly without regard for visual boundaries, many fall on class edges or ambiguous textures. Expanding these \textit{unsuitable} seed points generates regions that bleed across object boundaries, overwriting adjacent classes. While the original point label may be correct, the resulting pseudo-mask is not; it injects severe label noise that degrades any downstream segmentation model trained on it. As illustrated in Fig.~\ref{fig:intro}, our central hypothesis is that automatically distinguishing suitable from unsuitable points prior to propagation is critical for improving downstream segmentation performance in challenging real-world datasets.

While recent active learning approaches attempt to guide experts on \textit{where} to place new annotations for optimal model prompting~\cite{raine2024human,borja2026sseg}, we address a complementary problem: \textit{given the vast existing repositories of sparse annotations, which points can be trusted for propagation, and which parts of the resulting masks should be retained?} 

Overall, our contribution to  automated analysis of benthic imagery is twofold:%
\begin{itemize}
    \item \textbf{A Novel Framework for Point-Label Propagation.} We identify and quantify an overlooked failure mode in label propagation: the expansion of correctly-labeled but visually ambiguous points into harmful pseudo-masks. To solve this, we introduce an approach based on two novel stages:
    \begin{enumerate}
        \item \textbf{Pruning:} A pre-expansion stage that automatically identifies and discards \textit{unsuitable} points, preventing the generation of noisy masks.
        \item \textbf{Trimming:} A post-processing stage that refines the surviving masks by cutting and discarding regions that spill or overlap into adjacent classes. 
    \end{enumerate}
    Our experiments demonstrate that these combined stages significantly improve the quality of dense masks generated from sparse labels.
    
    \item \textbf{A New Benthic Segmentation  Benchmark.} We introduce a novel dataset derived from a long-term monitoring project, including expert point-labels placed at randomly selected locations. This dataset illustrates the real-world challenge of unsuitable propagation points. We validate our framework on this new data, as well as on established public datasets, demonstrating significant performance gains under randomly sampled point-labels.
\end{itemize}

To facilitate further research and provide the ecology community with improved analytical tools, our entire framework, including data, models, and code, is made publicly available. Our approach is directly extensible to other fine-grained ecological image datasets facing similar annotation constraints.


\section{Related Work}
\label{sec:related}

\paragraph{Sparse point-label propagation for benthic imagery.}
The CoralNet paradigm~\cite{beijbom2012automated} established sparse expert points as the
dominant annotation regime for benthic surveys. To train dense segmentation models from such labels,
CoralSeg~\cite{alonso2019coralseg} and PLAS~\cite{raine2022point} propagate points to
dense masks via superpixels, enforcing label purity within each superpixel;
D+NN~\cite{raine2024human} propagates with denoised DINOv2~\cite{oquab2024dinov2} features
and nearest-neighbor matching; and our own predecessor, SSeg~\cite{borja2026sseg}, expands points
with a hybrid SAM2-plus-superpixel scheme and actively guides where new points are placed.  However, often the labeled point set is already provided with the ecological surveys, and the active point-selection paradigm of SSeg can not be applied, unless a new labeling job is performed. 
CoralSCOP~\cite{zheng2024coralscop} is a coral-specialized foundation segmenter, an
alternative mask generator to SAM2. All of these treat the propagated region as trusted,
or filter only at the whole-superpixel level; none detects and cuts cross-class drift
\emph{within} a grown mask, nor calibrates a removal threshold against points that are likely-wrong or unsuitable for propagation for other reasons. This is precisely one of the ideas developed in our work, proposing a novel point pruning stage before propagating the point-labels into segments. 

Our pruning belongs to the instance-selection / label-noise filtering family and realizes the pruning principle of confident learning~\cite{northcutt2021confident}. 
Approaches such as AUM~\cite{pleiss2020aum} identify
mislabeled examples through training-time margin statistics; we obtain a \textit{verified} set of unsuitable points  directly from held-out reference points, so the threshold is calibrated without training or tuning any  model.

\paragraph{Pseudo-label reliability and refinement.}
In order to handle unreliable pseudo-labels we find semi-supervised methods based on confidence thresholding, such as 
FixMatch~\cite{sohn2020fixmatch} and its  adaptive successors
FlexMatch~\cite{zhang2021flexmatch} and FreeMatch~\cite{wang2023freematch}, or methods  treating
low-confidence regions separately U2PL~\cite{wang2022u2pl}. Mask-refinement methods such
as SAMRefiner~\cite{lin2025samrefiner} regrow cleaner masks with a foundation model. 
Our approach includes a refinement stage for trimming the obtained masks if necessary. It is inspired on the pseudo-mask refinement step in U2PL~\cite{wang2022u2pl}: 
we 
partition pixels into reliable and unreliable and keep the unreliable ones out of the computations. The main difference of our refinement step is how we compute the reliability signal. 
Ours is based on embedding distance to a per-image class prototype, instead of the previously used model confidence. A parallel line of work tolerates the noise at training time instead of filtering it. Noise-robust losses~\cite{zhang2018generalized,wang2019symmetric,ma2020normalized,ye2023active} bound the loss on mislabeled examples, assuming they produce a high loss. In our setting a mislabeled region is large, spatially coherent and internally consistent, so the model likely fits it fluently and at low loss, leaving nothing for a bounded loss to down-weight. We compare against Generalized Cross Entropy in Section~\ref{sec:main}

\paragraph{Prototype-based segmentation.}
Representing a class by the mean of its embeddings and classifying by nearest prototype is
a classic non-parametric approach~\cite{snell2017prototypical,wang2019panet}. 
Our trimming step is exactly a per-image nearest-prototype classifier in the style of
ProtoSeg~\cite{zhou2022protoseg} and PANet~\cite{wang2019panet}, but repurposed as a
reliability filter on a foundation-model mask rather than as the primary segmenter.

\paragraph{Foundation model for image segmentation.} 
We expand point-labels into dense segmentation masks by using each point as visual prompt for SAM2~\cite{ravi2024sam2,kirillov2023sam}, which returns the most likely segmentation for the object where that point is contained.  However, 
dense transformer features are known to carry
artifacts: high-norm outlier tokens~\cite{darcet2024registers} and position-driven grid patterns~\cite{yang2024denoising} that mislead similarity-based propagation. These artifacts can cause patch-level drift, which motivates checking the reliability of the propagated masks. Therefore, we use a complementary vision model to describe small image patches, 
DINOv3~\cite{simeoni2025dinov3}, which is the core ingredient in our verification steps, as detailed in Section~\ref{sec:method}.

\section{Benthic datasets: a new long-term monitoring benchmark}
\label{sec:dataset}

Among the existing public benthic datasets we select two commonly used in recent works that contain a large number of high quality semantic segmentation annotations for a large variety of classes. In particular, we use data from Coralscapes~\cite{coralscapes2025}, a densely-annotated coral-reef dataset of 39 classes whose images are extracted from underwater survey videos. The second is UCSD Mosaics~\cite{edwards2017large}, built from 16 large multi-species coral-reef mosaics with a resolution of $10\text{K}\times10\text{K}$ pixels and captured orthogonally, looking straight down at the seabed. CoralSeg~\cite{alonso2019coralseg} introduced a cropped version of this dataset by dividing the mosaics into $512\times512$ images, which we adopt in our work. Both of these datasets provide dense segmentation labels that allow us to evaluate the quality of the point-label propagation strategies considered in this work more accurately than with sparse point-labels frequently available in benthic surveys.

\begin{figure}[!tb]
    \centering
    \def\figheight{2.7cm} 

    \begin{subfigure}{0.31\linewidth}
        \centering
        \includegraphics[height=\figheight]{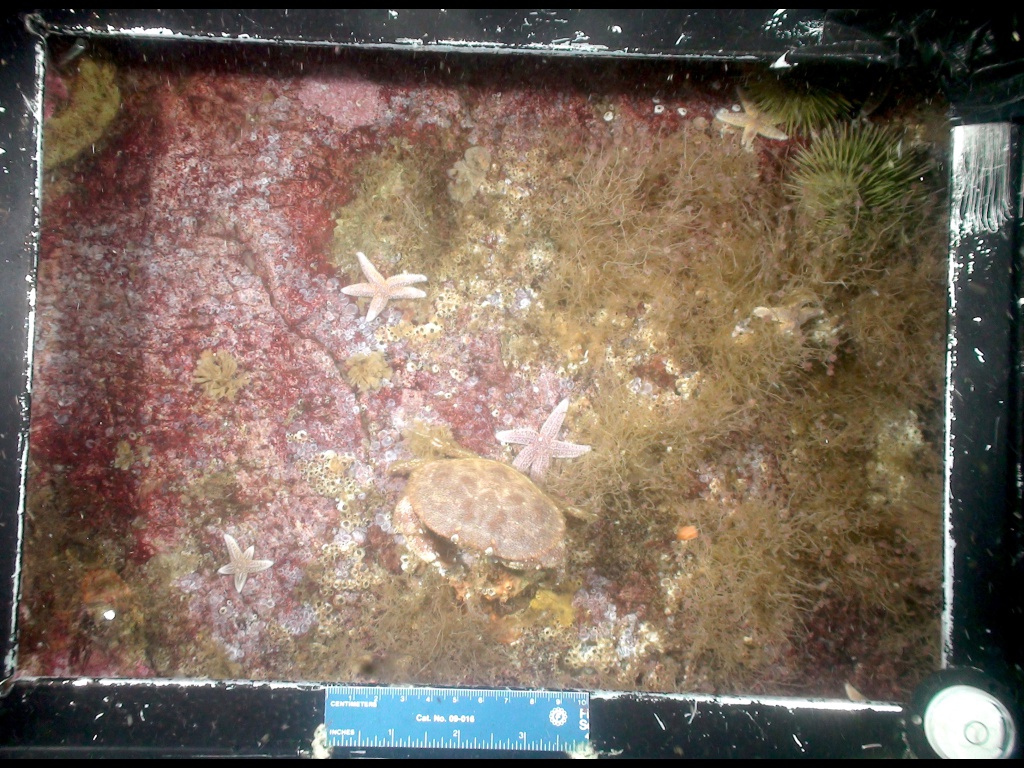}
        \caption*{\footnotesize (a) Our dataset}
    \end{subfigure}\hfill
    \begin{subfigure}{0.435\linewidth}
        \centering
        \includegraphics[height=\figheight]{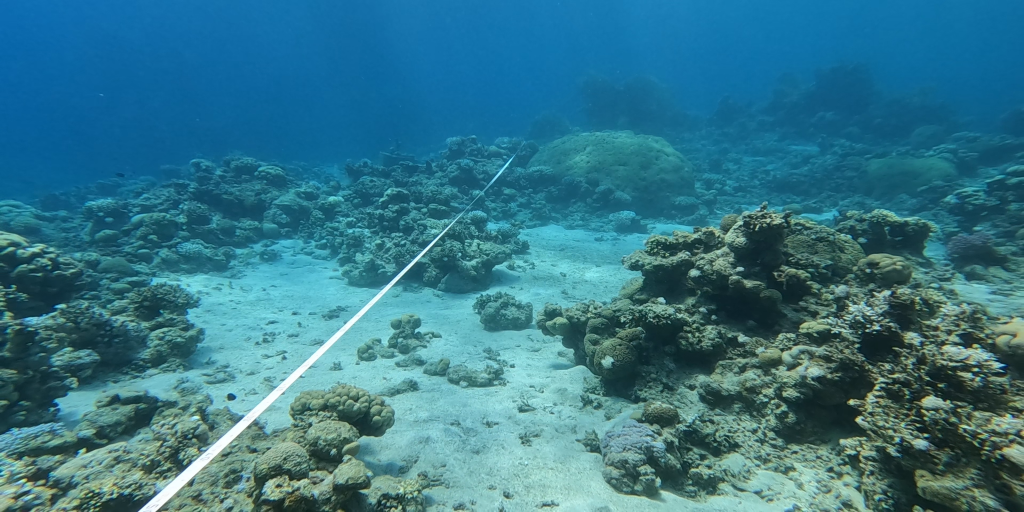}
        \caption*{\footnotesize (b) Coralscapes}
    \end{subfigure}\hfill
    \begin{subfigure}{0.25\linewidth}
        \centering
        \includegraphics[height=\figheight]{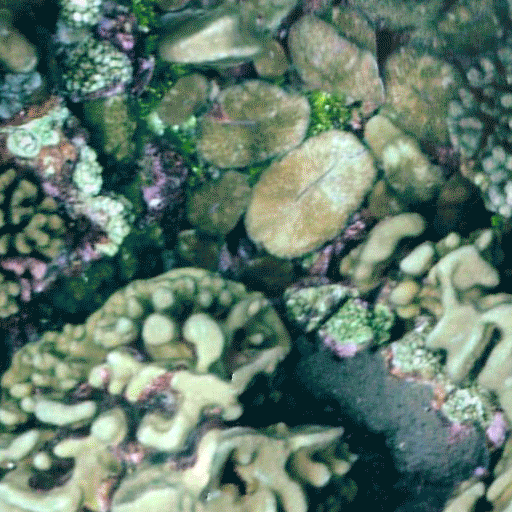}
        \caption*{\footnotesize (c) UCSD}
    \end{subfigure}
    \caption{Visual comparison of the benthic imagery datasets: (a) Ours, (b) CoralScapes~\cite{coralscapes2025}, and (c) UCSD Mosaics~\cite{edwards2017large, alonso2019coralseg}.}
    \label{fig:exampleDatasets}
\end{figure}

However, these datasets do not capture completely the challenges and limitations implicit to existing benthic imagery studies, such as those motivating this work. Therefore, we prepare and release a novel benchmark of data from a long-term monitoring project that includes images captured along large time spans (years). In addition, and actually what makes the tasks of label propagation more challenging, the point label annotation task was not explicitly guided by an end-task of segmentation. This means that the points were carefully annotated by marine biology experts, but these points were placed at random locations in the image.  Figure~\ref{fig:exampleDatasets} shows an example image from each of the three datasets.

The new presented benchmark
consists of benthic imagery covering 24 foreground classes, 313 training and 78 test images, and about 180 expert point-labels per image \cite{sebens2026halfwayrock}.  
The annotations performed by expert biologists consist of 200 random points placed on each photo using CoralNet. 
Points falling on mobile fauna were excluded from the analysis.

The train/test split is performed randomly at the image level. All the information of the data used in our experiments is available in the project site~\footnote{https://sites.google.com/unizar.es/benthic-seg}. 

The photos included in this benchmark represent a sample of all photos taken that need to be analyzed for a larger, novel long-term monitoring project examining how the sessile invertebrate community on subtidal rock walls has changed over time. This larger study represents one of very few ongoing temperate long-term subtidal monitoring projects thus providing a rich and unparalleled data source for documenting change in this diverse and dynamic ecosystem. The whole project consists of vast amounts of data, with photos taken annually in the summer at multiple sites around the same location. 
Therefore, developing automated segmentation techniques is an extremely helpful tool to allow the adequate  processing of all this survey data.

The class inventory in this dataset is very challenging for existing generic models and appearance-based methods, as it contains
near-identical colonial tunicate species, morphologically similar encrusting forms, and a
heterogeneous mixed-turf category.

\section{Segmentation-aware sparse point label propagation}
\label{sec:method}



\subsection{Overview}

\begin{figure}[!tb]
    
    \centering
    \includegraphics[width=\linewidth]{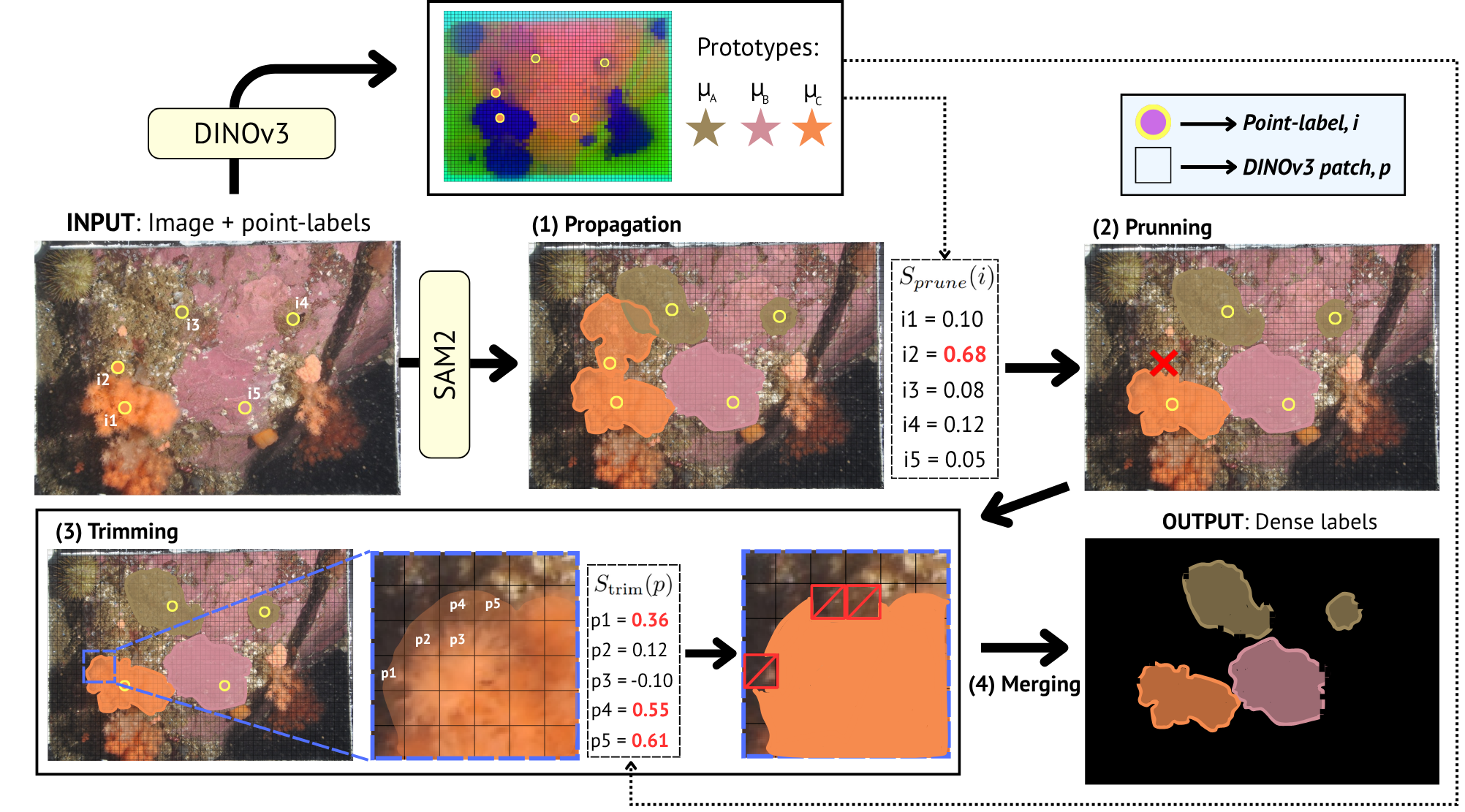}
    \vspace{-5mm}
    \caption{Overview of our pipeline. Given an image with sparse point-labels, we \textit{\textbf{propagate}} (1) each point-label $i$ with SAM2. Using DINOv3 features, we calculate per-image class prototypes $\mu_c$, and with them, we \textit{\textbf{prune}} (2) the point-labels whose mask leak into other classes ($i_2$), and \textit{\textbf{trim}} (3) the patches that disagree with their mask's class ($p1$, $p4$ and $p5$). The surviving masks are \textit{\textbf{merged}} (4) into the final dense labels by ground-truth voting.}
    \label{fig:overview}
\end{figure}

We propose a method that given an image with a set of sparse labeled points, generates a dense semantic segmentation of the image. In particular, each point of the fixed, pre-defined, set of sparse point-labels is propagated into a segment using SAM2 (i.e., each point is provided as a visual prompt for the model). The output of SAM2 is the most likely object segment where that point is contained. All the masks provided by SAM2 are combined into the final image segmentation. 
However, because SAM2 is class-agnostic, a point's specific spatial location can cause its expanded mask to bleed across semantic boundaries into regions of other classes. To detect and mitigate these semantic leaks, our method evaluates and refines the expansions through four sequential stages, as illustrated in Figure~\ref{fig:overview}:

\begin{enumerate}
    \item \textbf{Propagation}. Each given input labeled point is provided as a visual prompt for SAM2~\cite{ravi2024sam2}, in order to propagate the initial sparse point-labels, resulting in one independent dense mask per point-label.

\item \textbf{Pruning}.
With all the masks generated after propagating the input points, our pruning step identifies and removes point-labels, and their corresponding generated mask, that look unsuitable for propagation. Intuitively, this step filters out unreliable point-labels whose generated masks bleed into incorrect class regions. 
Section~\ref{sec:pruning} details how this proposed step is implemented.

\item  \textbf{Trimming}.
While pruning discards unreliable point-labels and their propagated masks entirely, the remaining masks may still contain localized errors. We evaluate each mask internally to identify regions that visually disagree with its class. Rather than discarding the mask completely, we \emph{trim} these inconsistent regions by excluding them from the mask, preserving only the reliable parts. Section~\ref{sec:trimming} describes this proposed process in more detail.

\item \textbf{Merging}. The trimmed, independent masks are combined into a single dense pseudo-mask, resolving any overlapping class conflicts via the same ground-truth voting proposed in SSeg~\cite{borja2026sseg}.

\end{enumerate}

\subsection{Pruning}
\label{sec:pruning}

This stage evaluates the initial set of sparse point-labels to identify and entirely remove those that are unsuitable for propagation with SAM2. To do this, we define a \emph{pruning score} for each point-label and a calibrated threshold to decide which points to remove along with its expansion. 

\paragraph{\textbf{Pruning Score}.} Each image is encoded once through DINOv3~\cite{simeoni2025dinov3} into a grid of patch embeddings. Every patch of $16\times16$ pixels is represented by a single $1024$-dimensional vector. Let $f_p$ be this vector for patch $p$, which we L2-normalize. Since the grid is fixed, every point-label lies in exactly one patch.

For every class $c$ present in the image, we define a local prototype $\mu_c$ as the L2-normalized mean of the embeddings from the patches that contain class-$c$ point-labels:%
\begin{equation}
    \mu_c = \frac{\sum_{p \in P_c} f_p}{\left\| \sum_{p \in P_c} f_p \right\|_2}, 
\end{equation}
\noindent where $P_c$ is that set of patches. Prototypes are per-image rather than global, so each one captures the appearance of its class under the illumination and blur of that specific scene.

Because $f_p$ and $\mu_c$ are unit vectors, their dot product equals their cosine similarity. Consider a point-label $i$ of class $y_i$ and its propagated mask $M_i$ from stage~(1). We look at the patches covered by $M_i$ and assign each one to the class $\hat{y}p$ of its nearest prototype:%
\begin{equation}
    \hat{y}p = \arg\max{c} (f_p \cdot \mu_c).
\end{equation}
\noindent 
Since $M_i$ is defined at the pixel level whereas patches are fixed $16\times16$ tiles, a patch at the mask boundary may fall only partially inside it. We count a patch as covered by $M_i$ when more than half of its pixels lie within the mask.

A patch $p$ \emph{disagrees} when $\hat{y}p \neq y_i$, i.e. it look more like another class than the one $i$ assigns to it. The pruning score $S_{prune}(i)$ is the fraction of the patches covered by $M_i$ that disagree:%
\begin{equation}
   S_{prune}(i) = \frac{\# \text{ patches of } M_i \text{ with } \hat{y}_p \neq y_i}{\# \text{ patches of } M_i}
\end{equation}
\noindent A high $S_{prune}(i)$ means that most of the region SAM2 would label $y_i$ actually looks like other classes, marking point-label $i$ as unsuitable for propagation.

\paragraph{\textbf{Verification-calibrated threshold}.} 

Rather than fixing the pruning threshold by hand, we calibrate it from the current data, as is common in label-error detection~\cite{northcutt2021confident}, against point-labels that we can \emph{verify} are unsuitable for propagation. To do so, we split each image's point annotations into an \emph{expansion set} (75\%), the points we actually propagate, and a \emph{validation set} (25\%) held out only as witnesses. A point-label of the expansion set is \emph{verified unsuitable} when its propagated mask $M_i$ covers a validation point of a different class: a direct proof, against a real held-out label, that its expansion leaks into another class.

We gather the pruning scores of all verified-unsuitable point-labels across the training images,%
\begin{equation}
    A_{\text{point}} = \{S_{\text{prune}}(i) \mid M_i \text{ covers a validation point-label of a different class}\},
\end{equation} 
\noindent and set the threshold to their median, $\tau_{\text{prune}} = \operatorname{median}(A_{\text{point}})$. A point-label $i$ is then pruned when $S_{\text{prune}}(i) \ge \tau_{\text{prune}}$, i.e. when it scores at least as poorly as a typical verified-unsuitable point. We use the median so the threshold reflects a \emph{typical} verified-unsuitable point rather than the extremes. If $A_{\text{point}}$ is empty, i.e. no propagated mask in the training set covers a validation point-label of a different class, no verified evidence of harmful propagation exists and we prune nothing, recovering the keep-all baseline. In practice this does not occur on the datasets we evaluate.

Finally, we standardize scores per image so a single threshold is
comparable across scenes. Points are then removed worst-first, in
decreasing order of $S_{prune}(i)$, subject to a class guard: a point is
pruned only while its class still retains more than one point-label in the
image, so every class keeps at least its lowest-scoring point-label. The
number of pruned points is thus not fixed but differs per image.

\subsection{Trimming}
\label{sec:trimming}

Pruning discards whole point-labels, but a surviving mask can still over-extend locally into regions of other classes. Trimming refines each surviving mask by removing such regions patch by patch, reusing the DINOv3 features and prototypes from the previous stage.

For a patch $p$ of a surviving mask of class $y_i$, we compare its cosine similarity to its own class prototype, $f_p \cdot \mu_{y_i}$, against its cosine similarity to the most similar class, $\max\limits_{c \neq y_i} (f_p \cdot \mu_c)$. The trimming score is the difference between the two:

\begin{equation}
    S_{\text{trim}}(p) = \max_{c \neq y_i} (f_p \cdot \mu_c) - (f_p \cdot \mu_{y_i})
\end{equation}
\noindent It is positive when the patch is more similar to another class than to its own, and grows the more it looks like that other class.

We calibrate the threshold as in pruning, now at patch granularity. We collect the trimming scores of the verified-unsuitable patches (those covering a validation point-label of a different class): %
\begin{equation}
    A_{\text{patch}} = \{ S_{\text{trim}}(p) \mid p \text{ covers a validation point-label of a different class} \},
\end{equation}
\noindent and take their median, $\tau_{\text{trim}} = \operatorname{median}(A_{\text{patch}})$. A patch $p$ is trimmed when $S_{\text{trim}}(p) \ge \tau_{\text{trim}}$, excluding all its pixels from the mask.

\section{Experiments}
\label{sec:experiments}

\subsection{Experimental setup}
\label{sec:setup}

\paragraph{\textbf{Semantic Segmentation network training.}}

To measure the quality of the generated pseudo-masks, we train a semantic segmentation model with them as supervision and evaluate the resulting model on the held-out test images. We train a SegFormer~\cite{xie2021segformer} architecture with a MiT-B2 encoder pretrained on ImageNet-1k. We optimize with SGD (momentum $0.9$, weight decay $0.01$) and layer-wise learning-rate decay (factor $0.9$): the encoder's top stage uses a learning rate of $7\times10^{-4}$, decayed by $0.9$ per earlier stage, and the decode head uses $10^{-3}$, with a $10\%$ linear warmup followed by cosine decay. We train with batch size $4$ at $512\times512$ resolution ($512\times1024$ for Coralscapes), for $160$ epochs on our dataset and $60$ on the larger Coralscapes and UCSD Mosaics, using standard augmentation (random horizontal flip, rotation up to $\pm30^\circ$, and color jitter).

\paragraph{\textbf{Datasets}}

We evaluate on the three datasets described in more detail in Sec.~\ref{sec:dataset}: our benthic dataset (24 classes), Coralscapes~\cite{coralscapes2025} (39 classes), and UCSD Mosaics~\cite{edwards2017large, alonso2019coralseg} (34 classes).  Our dataset provides a mean of 180 expert point-labels per image, over 313 training and 78 test images. Coralscapes and UCSD Mosaics are densely annotated, so we randomly sample 180 sparse point-labels per image from their training masks, matching the mean point density of \ourdata{}, to reproduce the weakly-supervised setting. We use the available train sets of 1683 and 3974 images, respectively, and evaluate on their corresponding test sets of 392 and 696 images.

\paragraph{\textbf{Metrics}}
We report standard mean Intersection-over-Union (mIoU) and mean per-class accuracy (mPA) over all pixels for 
Coralscapes and UCSD Mosaics, since they provide dense ground truth masks for test image. Since our dataset does only provide sparse point labels, we can only report mean per-class accuracy (mPA) evaluated on the available point annotations. Note that although mIoU is a more robust indicator for segmentation tasks, the trends observed are the same with both metrics. 
We exclude the background classes excluded from the mean in both cases.

\subsection{Segmentation results with different propagation strategies}
\label{sec:main}

 Our main evaluation compares SegFormer performance when trained with SAM2-propagated pseudo-masks under different strategies for handling unreliable propagated supervision. All strategies receive the same point-labels and identical supervision, so the comparison isolates the effect of the filtering strategy and reports the \emph{relative} ranking of strategies in a weakly-supervised sparse-point regime, not absolute performance comparable to densely-supervised training. Table~\ref{tab:main} summarizes the results of this experiment. Our strategy produces the largest gains on the two datasets with noisier labels, where it removes $34\%$ (\ourdata{}) and $25\%$ (Coralscapes) of the points.

The three datasets behave differently under filtering. On \ourdata{} and
Coralscapes, where labels are noisier, Ours clearly separates from
\textit{keep all}. UCSD Mosaics is a special case, where all methods
instead cluster around $44$~mIoU. Our hypothesis is that it presents a
situation less prone to unsuitable point propagations, with high quality
images and dense ground-truth masks with sharp boundaries, where sampled
points rarely fall on ambiguous regions. The emergent drop rate supports
this: our method removes only $11.7\%$ of the point-labels on UCSD
Mosaics, against $24.6\%$ on Coralscapes, indicating cleaner labels with
fewer unsuitable propagations. Here, pruning is not relevant, but we do
not penalize the baseline 44.41 mIoU as much as with other approaches to
tackle noise (GCE).

The random baselines drop as many points as Ours, but at random. On the two noisier datasets they do not beat \textit{keep all}, and on \ourdata{} they score well below it ($22.6$ and $23.5$ vs.\ $27.5$). This shows how pruning point-labels does not help by itself, what matters is \emph{which} points are pruned. GCE keeps every label and down-weights the noisy ones in the loss, but it only matches mPA obtained with the baseline \textit{keep all} on all three datasets, and even lowers at UCSD mIoU, showing that our pruning strategy results more effective.

\begin{table}[t]
\centering
\caption{Filtering propagated labels by reliability across three datasets. All strategies propagate the same expert point-labels into dense pseudo-masks with SAM2 and train Segformer on them. \textbf{\textit{Keep all}} trusts every propagated label; \textbf{GCE} also keeps all labels but trains with a noise-robust loss; the random baselines discard labels without any reliability estimate; \textbf{Ours} prunes unreliable points before propagation and trims unreliable regions afterwards. Our benthic dataset has no dense annotation and is evaluated only at its expert-labeled points. Mean over 5 seeds for \ourdata{}, a single seed for Coralscapes and UCSD Mosaics.}

\label{tab:main}
\resizebox{0.99\textwidth}{!}{%
\begin{tabular}{lcccccccc}
\toprule
 & \multicolumn{2}{c}{Our dataset} & \multicolumn{3}{c}{Coralscapes~\cite{coralscapes2025}} & \multicolumn{3}{c}{UCSD Mosaics~\cite{edwards2017large, alonso2019coralseg}} \\
\cmidrule(lr){2-3} \cmidrule(lr){4-6} \cmidrule(lr){7-9}
Strategy & mPA$^{*}$ & used\% & mPA & mIoU & used\% & mPA & mIoU & used\% \\
\midrule
None (\textit{keep all})                  & $27.53$ & 100 & 33.71 & 25.57 & 100 & 53.75 & 44.41 & 100 \\
GCE~\cite{zhang2018generalized}   & $27.86$ & 100 & 33.54 & 25.52 & 100 & 53.87 & 41.66 & 100 \\
\midrule
Blind random drop \textdagger         & $22.60$ & 66.1 & 33.76 & 25.39 & 75.4 & \textbf{54.77} & \textbf{45.12} & 88.3 \\
Class-balanced random \textdagger    & $23.46$ & 66.1 & 33.95 & 25.79 & 75.4 & 52.11 & 42.73 & 88.3 \\
\textbf{Ours}             & $\mathbf{32.00}$ & 66.1 & \textbf{37.21} & \textbf{27.60} & 75.4 & 53.60 & 44.05 & 88.3\\
\bottomrule
\multicolumn{9}{l}{\footnotesize $^{*}$Only evaluated at the GT points. \textdagger Dropped points \% matched to Ours.}\\
\end{tabular}
}
\end{table}

\subsection{Pruning and Trimming ablation}
This ablation experiment, summarized in  
Table~\ref{tab:ablation_cleaning}, isolates \emph{pruning} and \emph{trimming} stages, explained in Section~\ref{sec:method}, on \ourdata{}. Each of them is shown to contribute to improving the final result: trimming raises mPA from $27.5$ to $29.8$ and pruning to $31.0$, suggesting that the significance of pruning bad points before propagation is higher than trimming bad regions afterwards, but both result complementary. 
\begin{table}[h]
    \centering
    \caption{Pruning and trimming ablation on \ourdata{}.}
    \label{tab:ablation_cleaning}
    \begin{tabular}{lc}
        \toprule
        \textbf{Method} & \textbf{mPA} \\
        \midrule
        keep all  & $27.53 \pm 1.0$ \\
        trim only & $29.82 \pm 0.9$ \\
        prune only & $31.00 \pm 0.7$ \\
        prune + trim (ours) & $\mathbf{32.00 \pm 1.1}$ \\
        \bottomrule
    \end{tabular}
\end{table}

\subsection{Qualitative results}

\begin{figure*}[!thb]
    \centering
    \begin{subfigure}{0.32\linewidth}
        \includegraphics[width=\linewidth]{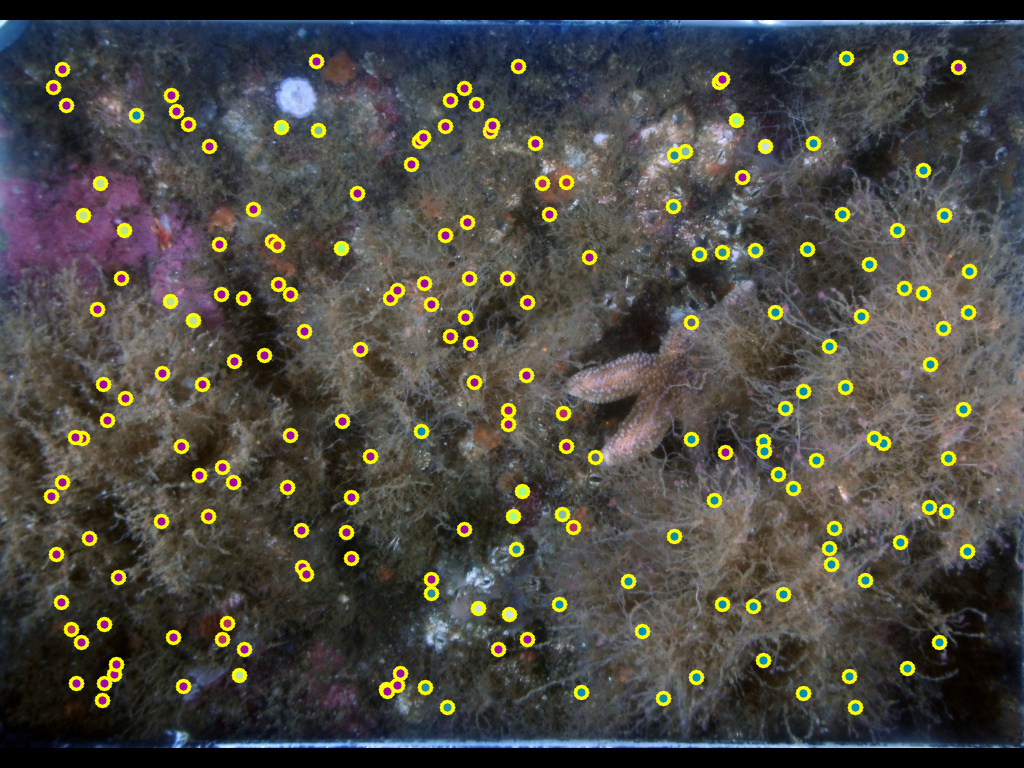}
    \end{subfigure}
    \begin{subfigure}{0.32\linewidth}
        \includegraphics[width=\linewidth]{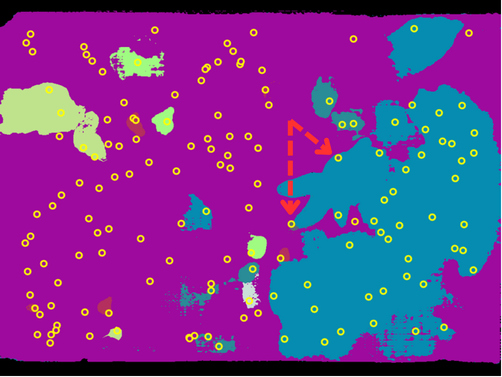}
    \end{subfigure}
    \begin{subfigure}{0.32\linewidth}
        \includegraphics[width=\linewidth]{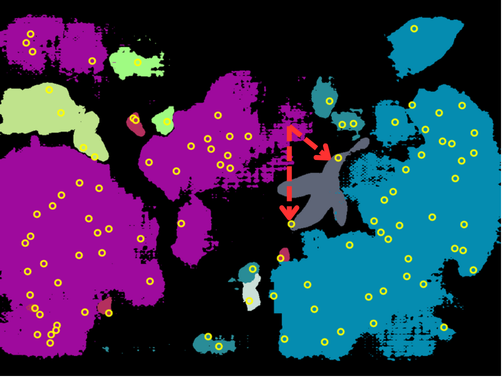}
    \end{subfigure}

    \vspace{2mm}

    \begin{subfigure}{0.32\linewidth}
        \includegraphics[width=\linewidth]{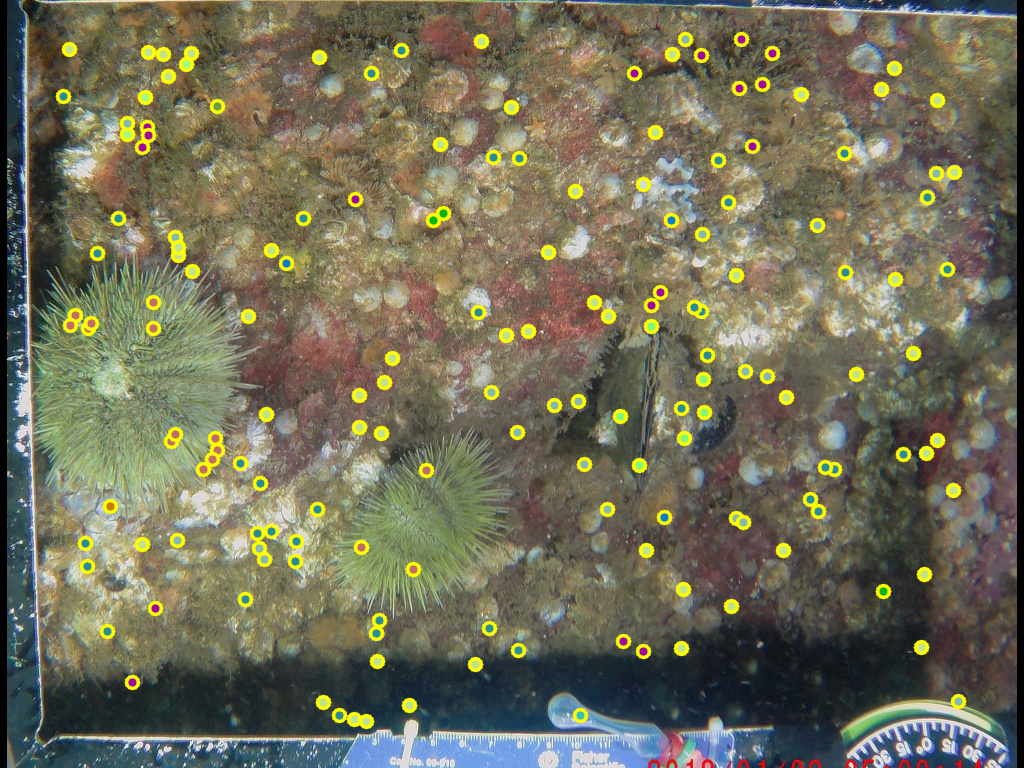}
    \end{subfigure}
    \begin{subfigure}{0.32\linewidth}
        \includegraphics[width=\linewidth]{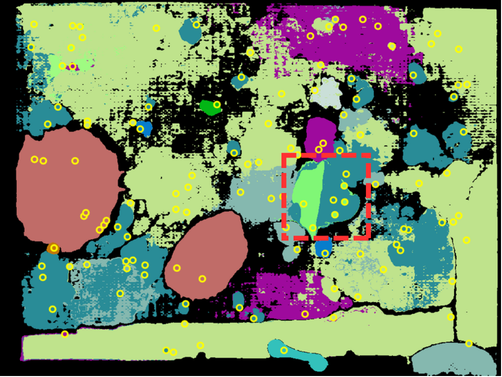}
    \end{subfigure}
    \begin{subfigure}{0.32\linewidth}
        \includegraphics[width=\linewidth]{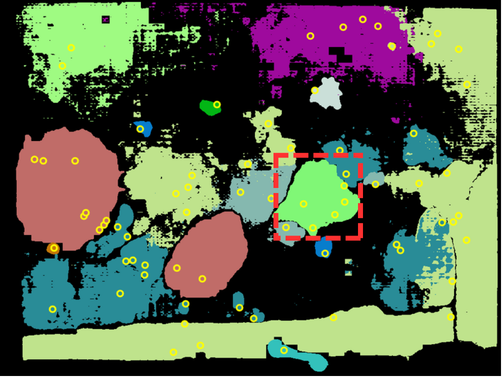}
    \end{subfigure}

    \vspace{2mm}

    \begin{subfigure}{0.32\linewidth}
        \includegraphics[width=\linewidth]{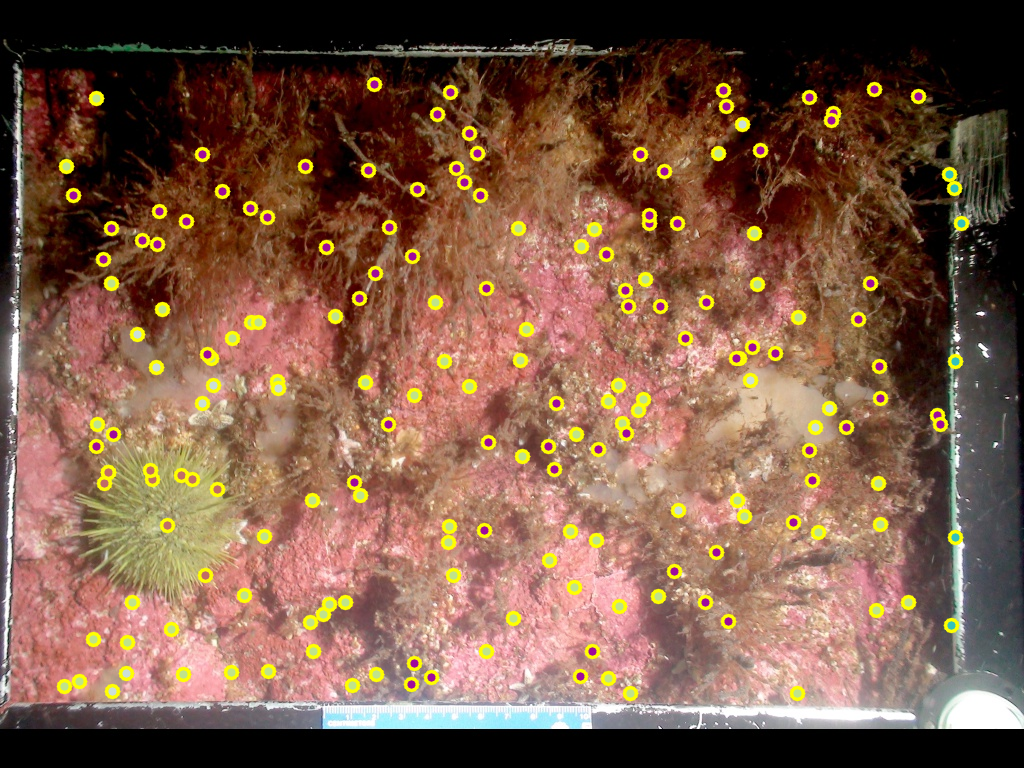}
    \end{subfigure}
    \begin{subfigure}{0.32\linewidth}
        \includegraphics[width=\linewidth]{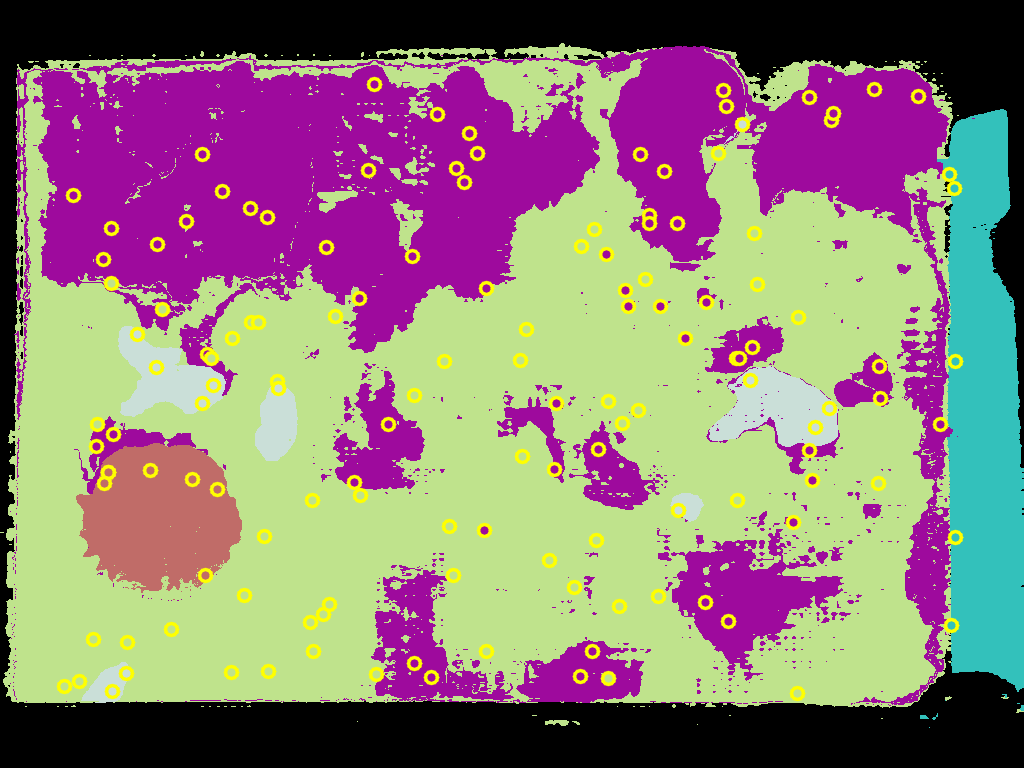}
    \end{subfigure}
    \begin{subfigure}{0.32\linewidth}
        \includegraphics[width=\linewidth]{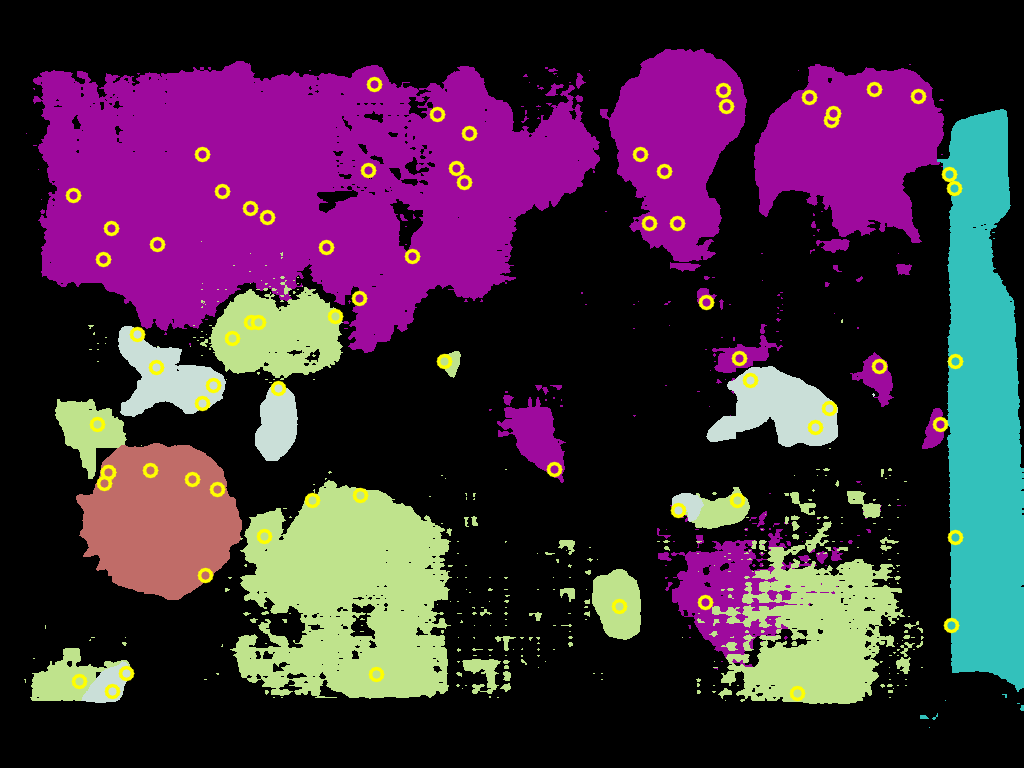}
    \end{subfigure}


    \begin{subfigure}{0.32\linewidth}
        \includegraphics[width=\linewidth]{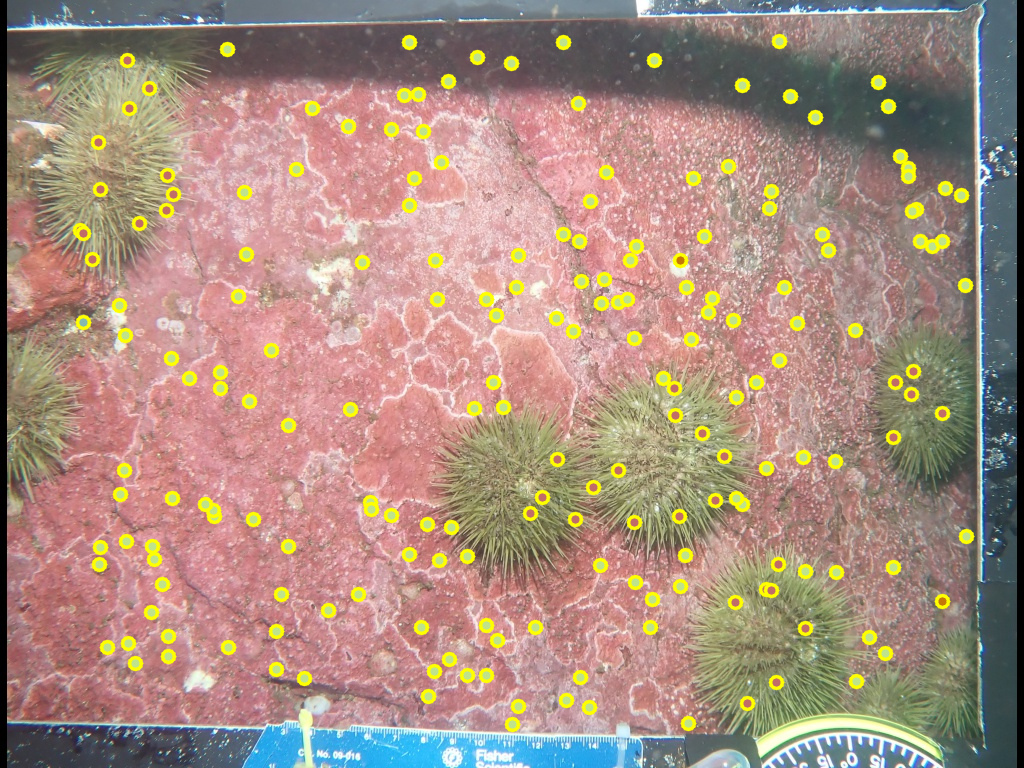}
        \caption*{\footnotesize (a) Image}
    \end{subfigure}
    \begin{subfigure}{0.32\linewidth}
        \includegraphics[width=\linewidth]{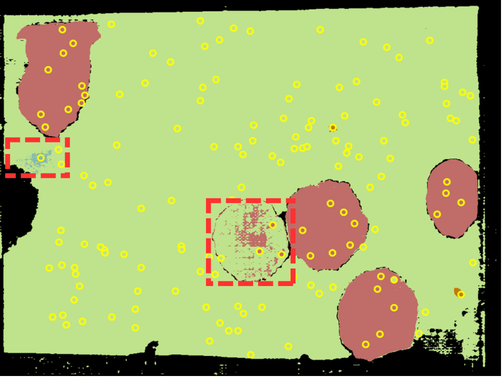}
        \caption*{\footnotesize (b) Keep all}
    \end{subfigure}
    \begin{subfigure}{0.32\linewidth}
        \includegraphics[width=\linewidth]{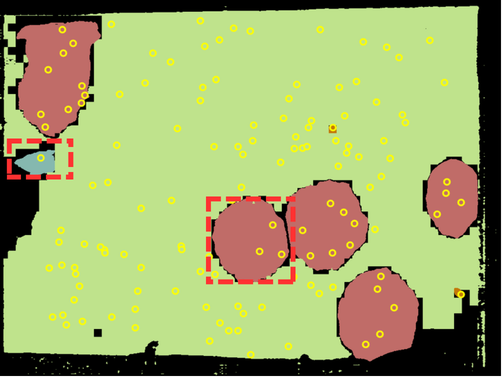}
        \caption*{\footnotesize (c) Ours}
    \end{subfigure}
    \vspace{2mm}
    
    \caption{Qualitative examples on \ourdata{}. Each row shows, (a) the input image with the available the point-labels overlaid as class-colored dots (with yellow edge for visibility); (b) the masks obtained after propagating all points; (c) the masks obtained with our approach. (a) and (b) show all the point-labels while (c) only shows the points that survive our \textit{pruning}. All examples show situations where expert correct point-labels get incorrectly covered by other class propagation. Red dashed boxes and arrows highlight some of the interesting cases. Pruning discards points whose expansion is unreliable and trimming removes mislabeled regions. In some cases removing an over-extended mask is key to enable neighbouring mask supply the correct label, recovering pixels that keep-all had mislabeled (e.g., the example in row 1).}
\label{fig:ours_qualitative}
\end{figure*}

\begin{sidewaysfigure}
    \centering
    \begin{subfigure}{0.24\linewidth}
        \includegraphics[width=\linewidth]{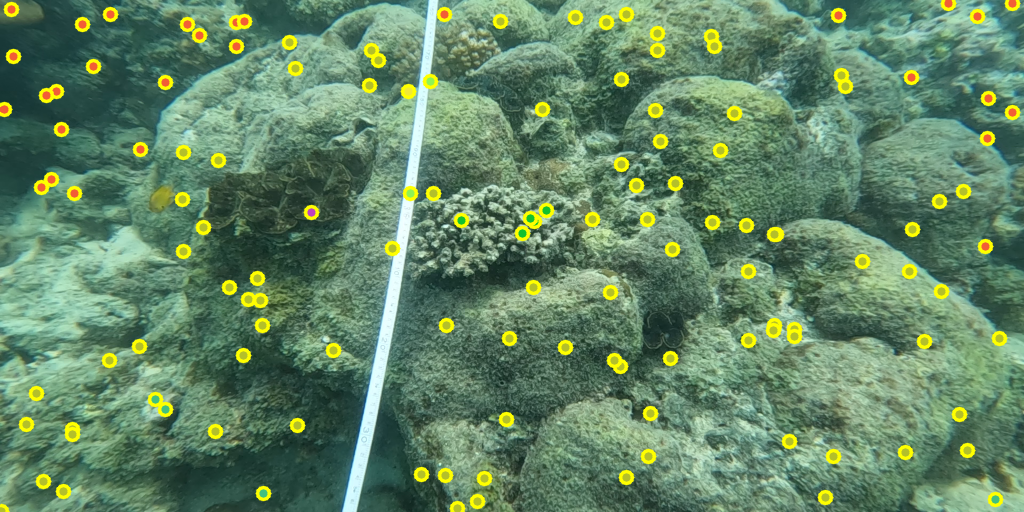}
    \end{subfigure}
    \begin{subfigure}{0.24\linewidth}
        \includegraphics[width=\linewidth]{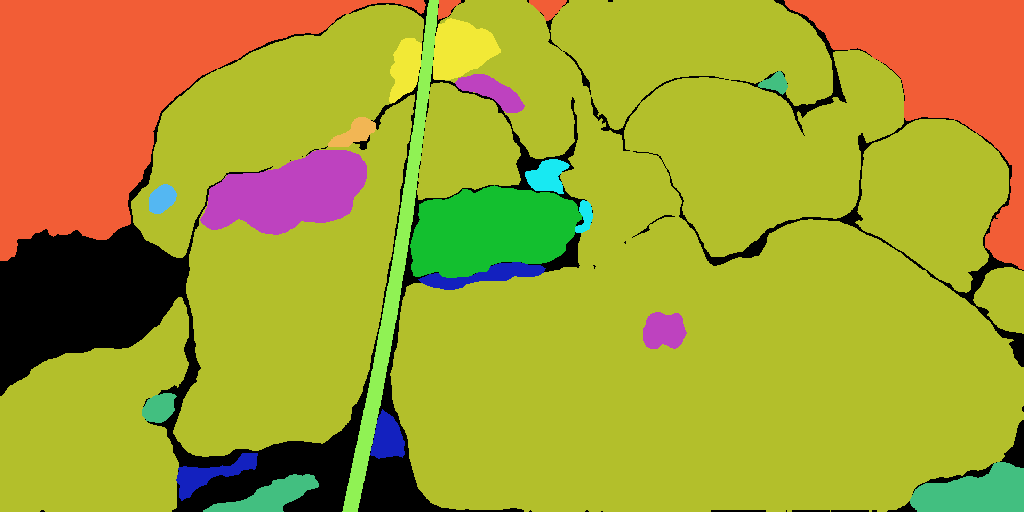}
    \end{subfigure}
    \begin{subfigure}{0.24\linewidth}
        \includegraphics[width=\linewidth]{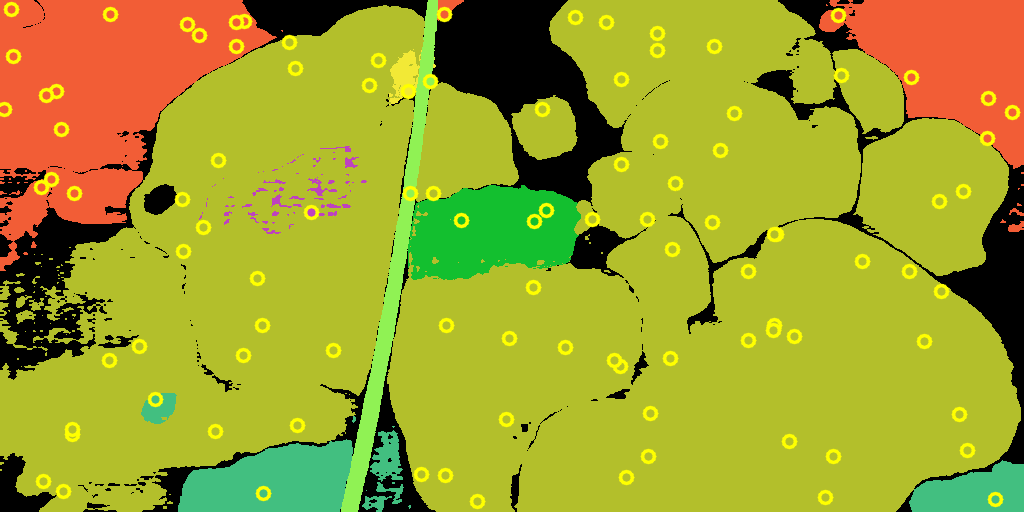}
    \end{subfigure}
    \begin{subfigure}{0.24\linewidth}
        \includegraphics[width=\linewidth]{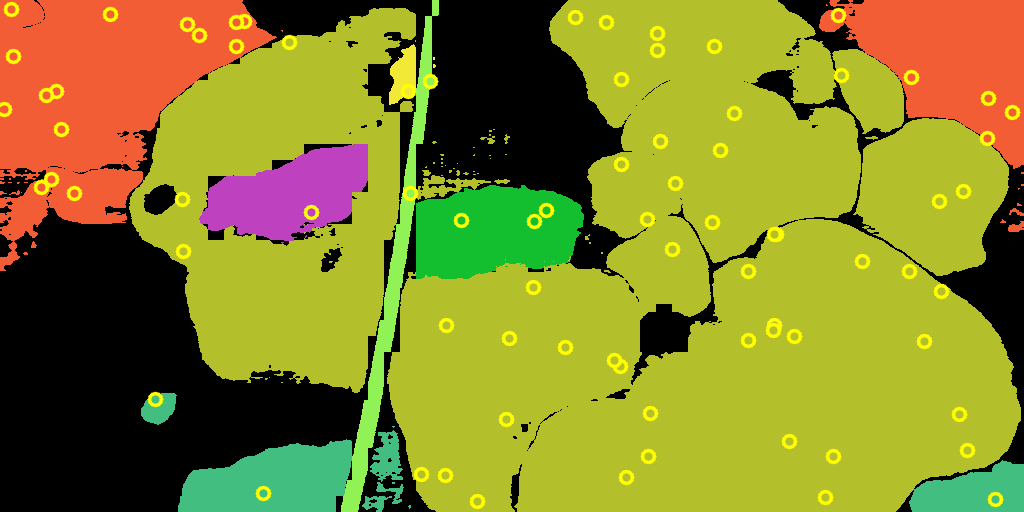}
    \end{subfigure}
    \vspace{2mm}
    \begin{subfigure}{0.24\linewidth}
        \includegraphics[width=\linewidth]{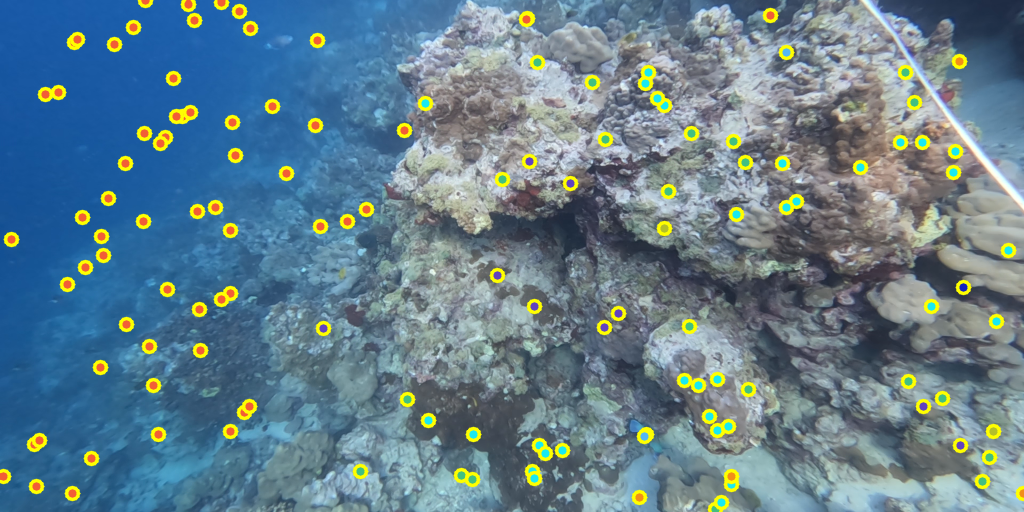}
    \end{subfigure}
    \begin{subfigure}{0.24\linewidth}
        \includegraphics[width=\linewidth]{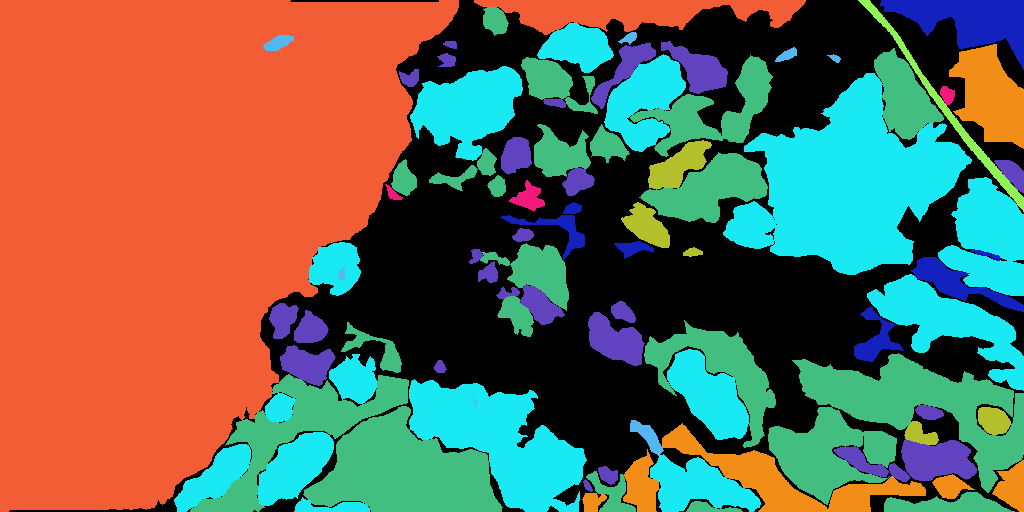}
    \end{subfigure}
    \begin{subfigure}{0.24\linewidth}
        \includegraphics[width=\linewidth]{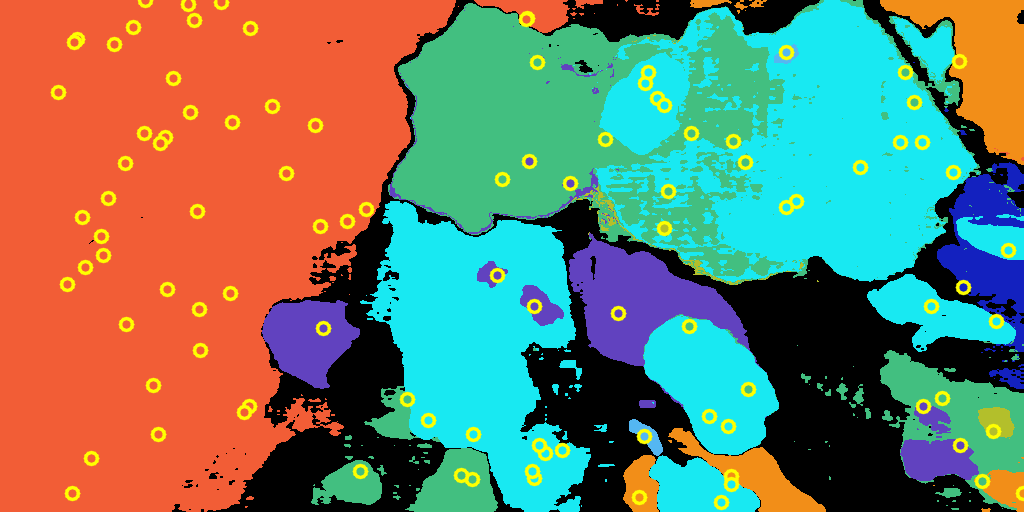}
    \end{subfigure}
    \begin{subfigure}{0.24\linewidth}
        \includegraphics[width=\linewidth]{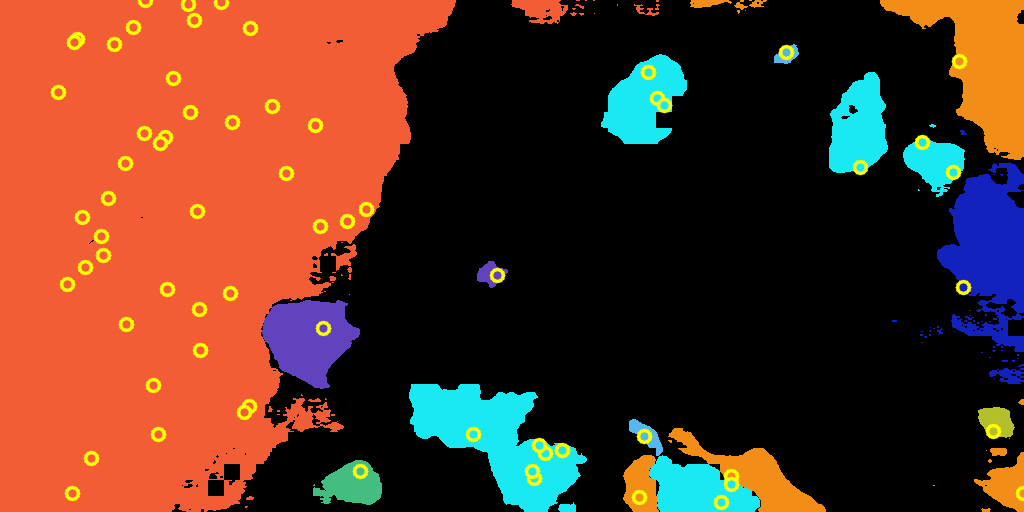}
    \end{subfigure}
    \vspace{2mm}
    \begin{subfigure}{0.24\linewidth}
        \includegraphics[width=\linewidth]{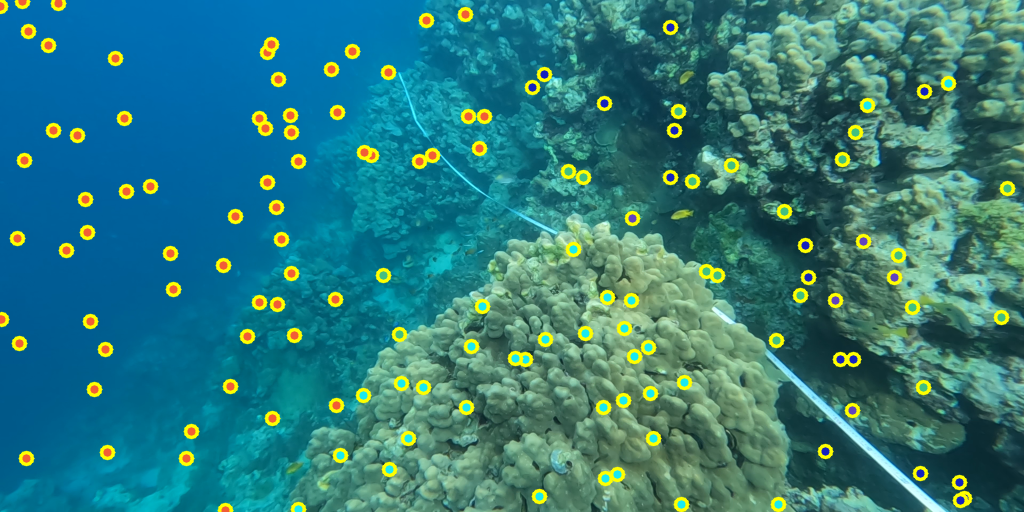}
    \end{subfigure}
    \begin{subfigure}{0.24\linewidth}
        \includegraphics[width=\linewidth]{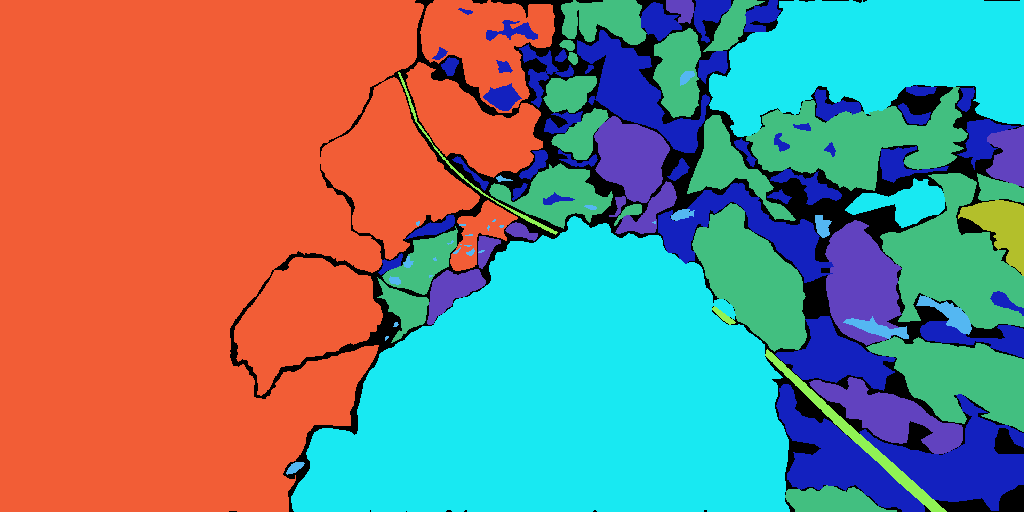}
    \end{subfigure}
    \begin{subfigure}{0.24\linewidth}
        \includegraphics[width=\linewidth]{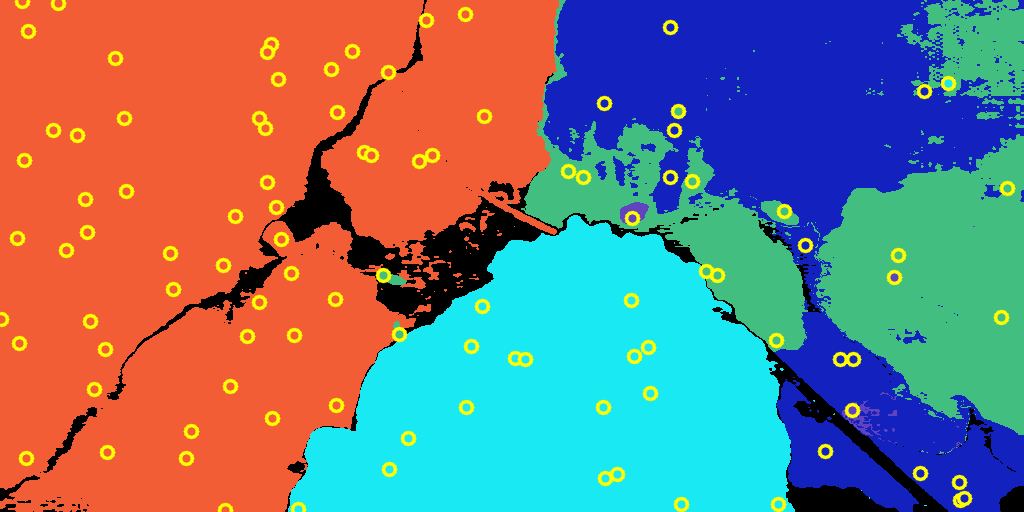}
    \end{subfigure}
    \begin{subfigure}{0.24\linewidth}
        \includegraphics[width=\linewidth]{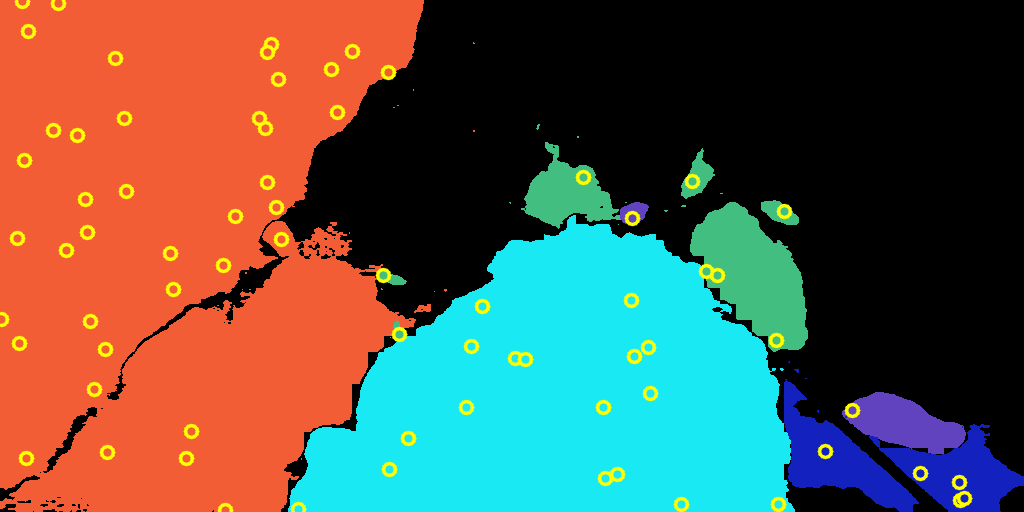}
    \end{subfigure}
    \vspace{2mm}
    \begin{subfigure}{0.24\linewidth}
        \includegraphics[width=\linewidth]{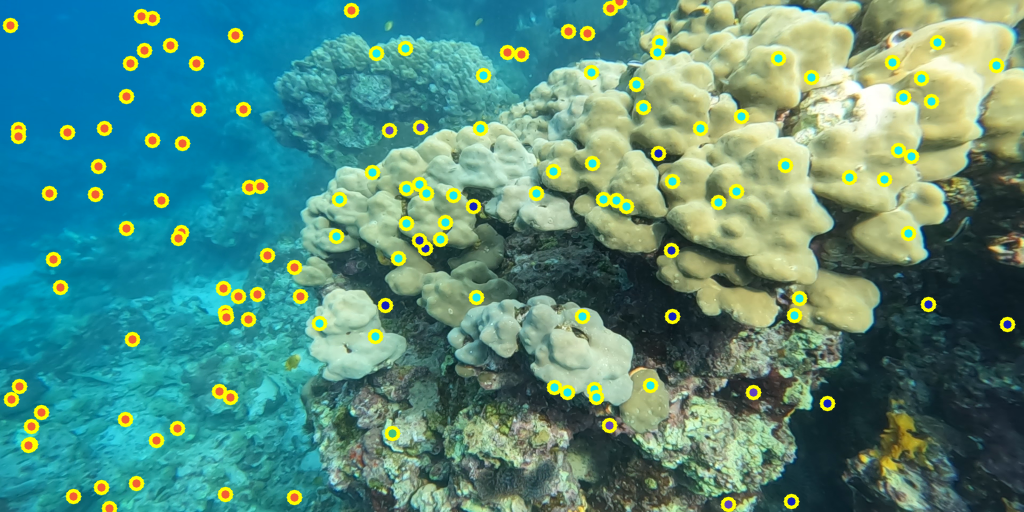}
        \caption*{\footnotesize (a) Image}
    \end{subfigure}
    \begin{subfigure}{0.24\linewidth}
        \includegraphics[width=\linewidth]{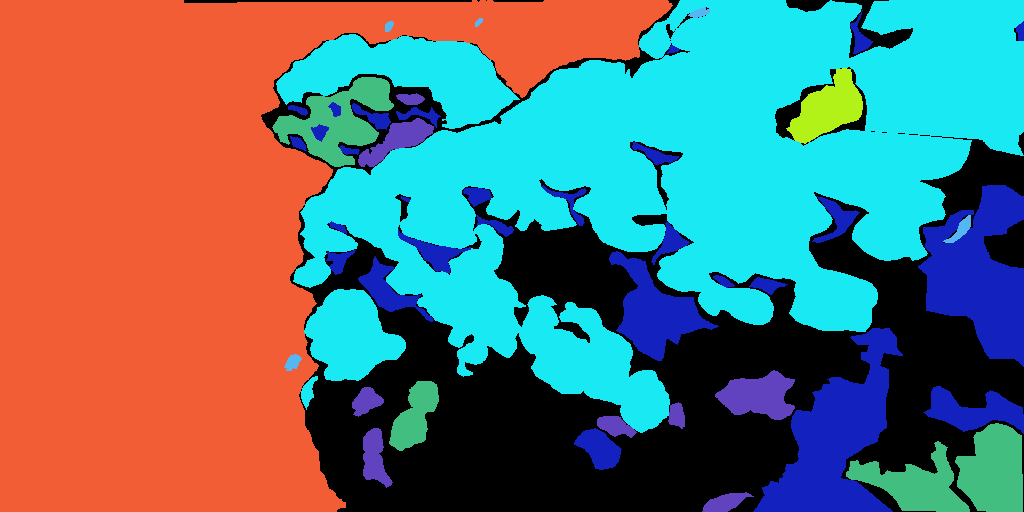}
        \caption*{\footnotesize (b) Ground truth}
    \end{subfigure}
    \begin{subfigure}{0.24\linewidth}
        \includegraphics[width=\linewidth]{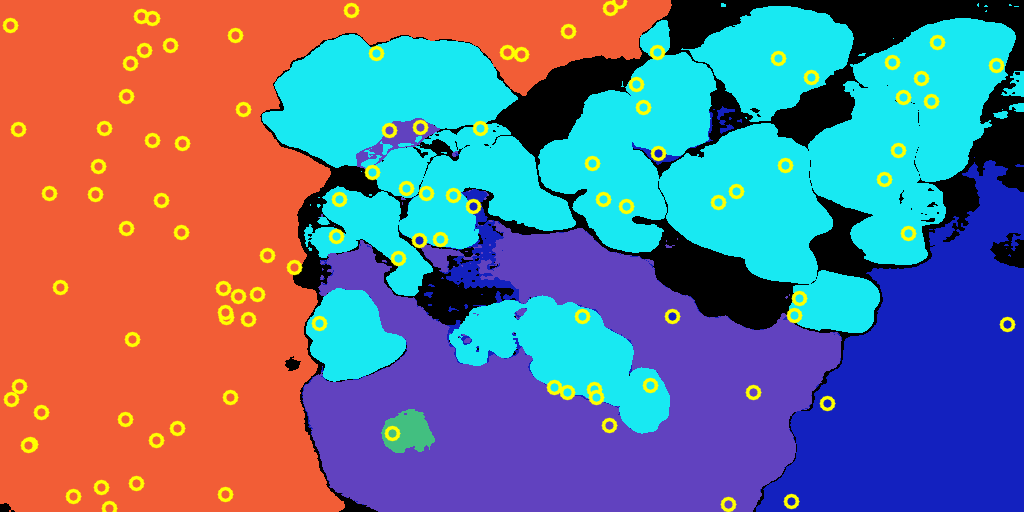}
        \caption*{\footnotesize (c) Keep all}
    \end{subfigure}
    \begin{subfigure}{0.24\linewidth}
        \includegraphics[width=\linewidth]{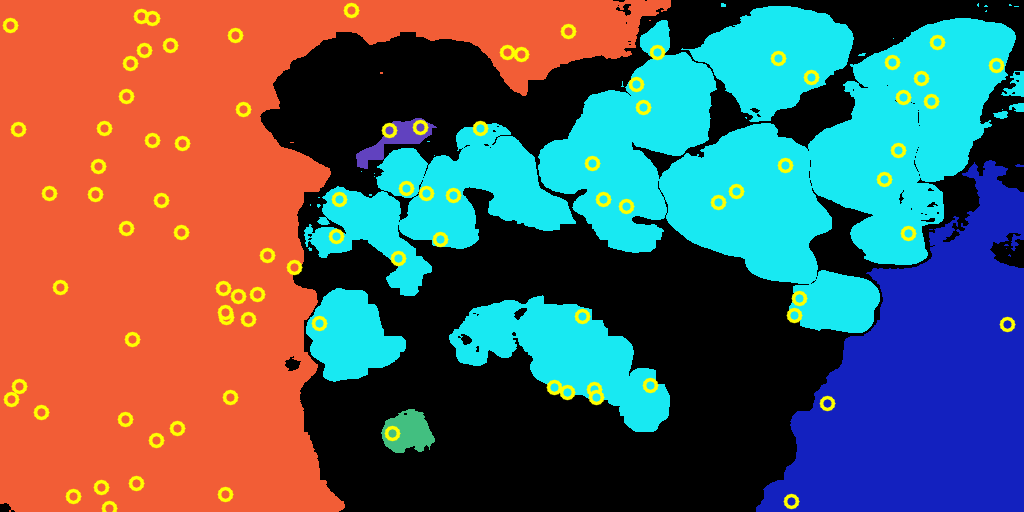}
        \caption*{\footnotesize (d) Ours}
    \end{subfigure}
    \vspace{2mm}
    
    \caption{Qualitative examples on Coralscapes. Each row shows, (a) the input image with the available the point-labels overlaid as class-colored dots (with yellow edge for visibility); (b) the Ground truth; (c) the masks obtained after propagating all points; (d) the masks obtained with our approach. (a) and (c) show all the point-labels while (d) only shows the points that survive our \textit{pruning}. These examples show two situations our method corrects: expert-correct point-labels whose SAM2 mask bleeds into a neighbouring class, and correct labels whose mask over-expands far beyond the object. Pruning discards the point when its whole expansion is unreliable, and trimming removes only the regions of a surviving mask that disagree with its class. As a result, \textit{ours} often leaves a region unlabelled rather than keep the confidently-wrong, over-extended mask that keep-all propagates.}
\label{fig:coralscapes_qualitative}
\end{sidewaysfigure}

We complement the quantitative results with qualitative examples on \ourdata{} and Coralscapes. Figures~\ref{fig:ours_qualitative} and ~\ref{fig:coralscapes_qualitative} compares, for several images, the pseudo-mask obtained by propagating all points against \textit{Ours}, with the point-labels overlaid, showing how pruning and trimming remove unreliable expansions and mislabeled regions and, in some cases, recover the correct label for pixels that keep-all had mislabeled.

\section{Conclusion}
\label{sec:conclusion}

In this work, we addressed a critical bottleneck to enable  scaling automated analysis of marine ecological monitoring data, by bridging the gap between sparse legacy annotations and modern dense semantic segmentation models. While visual foundation models like SAM2 offer immense potential for automated image analysis, we demonstrated that their direct application to randomly sampled survey points frequently introduces harmful label noise due to boundary ambiguity. To overcome this, we introduced a novel point-label propagation framework that explicitly accounts for the suitability of visual prompts before and after expansion. 

By integrating a pre-processing \textit{pruning} stage to filter out unsuitable points for propagation, and a post-processing \textit{trimming} stage to resolve class overlaps, our method significantly enhances the quality of generated pseudo-ground-truth masks. We validated these contributions on established public benchmarks and introduced a new challenging dataset featuring real-world sparse expert annotations. Our experiments confirm that automatically identifying and discarding unsuitable points mitigates the degradation of downstream segmentation models, leading to notable performance gains in fine-grained benthic semantic analysis.

These gains also come with limitations. The appearance-based signals rely on per-image class prototypes, so a class with few point-labels yields an unreliable prototype. Near-identical classes also remain hard to separate when appearance alone is not enough. Even so, this framework provides a robust and practical tool to unlock the value of vast, existing ecological image repositories without requiring additional, expensive human annotation effort. By releasing our code, models, and the new benchmark dataset, we aim to facilitate ongoing interdisciplinary collaboration between the computer vision and marine biology communities. Future work will explore extending these filtering mechanisms to other ecological domains with similar annotation constraints, further assisting on the computation of valuable metrics for ecological studies from complex environmental data.

\section*{Acknowledgments}
This work was supported by grants AIA2025-163563-C31 and PID2024-159284NB-I00, funded by 
MICIU/AEI /10.13039/501100011033 and ERDF$\backslash$EU, and by DGA project T45\_23R. BAM and JEKB were partially supported by the Stone Living Lab and Woods Hole Sea Grant while working on data for this manuscript.



\bibliographystyle{splncs04}
\bibliography{main}

\end{document}